\documentclass[12pt,a4paper]{article}
\usepackage{lineno}
\usepackage{capt-of}
\usepackage[utf8]{inputenc}
\usepackage{setspace}
\usepackage{caption}
\usepackage{hyperref}
\usepackage{tabularx}
\usepackage{tcolorbox}
\usepackage{pdflscape}
\usepackage{mathtools}
\usepackage{pdfpages}
\usepackage{setspace}
\usepackage{inputenc}
\usepackage{enumitem}
\setlist[itemize]{noitemsep, nolistsep, left=0pt}
\usepackage{float}
\usepackage{listings}
\usepackage{amssymb, amsmath}
\hypersetup{
    colorlinks=true,
    linkcolor=blue,
    filecolor=magenta,
    citecolor=black,
    urlcolor=cyan,
    pdftitle={Overleaf Example},
    pdfpagemode=FullScreen,
    }
\usepackage{authblk}
\usepackage{footnote}
\usepackage{comment}
\usepackage{authblk}
\usepackage{listings}

\usepackage{import}
\usepackage{xcolor}
\usepackage{soul}

\usepackage{array}
\usepackage{longtable,booktabs,threeparttablex}

\usepackage[
    backend=biber,
    style=apa,
  ]{biblatex}
\usepackage{csquotes}

\hypersetup{breaklinks=true} % Enable line breaks in links
\begin{document}

\title{Detecting Religious Language in Climate Discourse}
\author{Evy Beijen\footnote{Eep Talstra Centre for Bible and Computer, Vrije Universiteit Amsterdam, email: \url{evybeijen@gmail.com}}, Pien Pieterse\footnote{Eep Talstra Centre for Bible and Computer, Vrije Universiteit Amsterdam, email: \url{pienpieterse@gmail.com}}, Yusuf Çelik\footnote{Eep Talstra Centre for Bible and Computer, Vrije Universiteit Amsterdam, email: \url{y.celik@vu.nl}}, Willem Th. van Peursen\footnote{Eep Talstra Centre for Bible and Computer, Vrije Universiteit Amsterdam, email: \url{w.t.van.peursen@vu.nl}},  \\
Sandjai Bhulai\footnote{Department of Mathematics, Vrije Universiteit Amsterdam, email: \url{s.bhulai@vu.nl}},  Meike Morren\footnote{Department of Marketing, School of Business and Economics, Vrije Universiteit Amsterdam, email: \url{meike.morren@vu.nl}}}

% \maketit

\date{\today}

\maketitle
\begin{abstract}
Religious language continues to permeate contemporary discourse, even in ostensibly secular domains such as environmental activism and climate change debates. This paper investigates how explicit and implicit forms of religious language appear in climate-related texts produced by secular and religious nongovernmental organizations (NGOs). We introduce a dual methodological approach: a rule-based model using a hierarchical tree of religious terms derived from ecotheology literature, and large language models (LLMs) operating in a zero-shot setting. Using a dataset of more than 880,000 sentences, we compare how these methods detect religious language and analyze points of agreement and divergence. The results show that the rule-based method consistently labels more sentences as religious than LLMs. These findings highlight not only the methodological challenges of computationally detecting religious language but also the broader tension over whether religious language should be defined by vocabulary alone or by contextual meaning. This study contributes to digital methods in religious studies by demonstrating both the potential and the limitations of approaches for analyzing how the sacred persists in climate discourse.
\end{abstract}

\section{Introduction} \label{introduction}
Religion has been an integral part of the human experience for millennia. Although we may not know the full extent of its influence, material evidence and historical records make it clear that, in Europe, a wide range of religious traditions have emerged, flourished, and faded over the centuries. The ritual practices of Germanic paganism, the state-sanctioned worship of the Roman imperial cult, and the sweeping transformation brought about by Christianity are among the many traditions that have shaped European history. In light of this history, it is hardly surprising that the very foundations of Indo-European languages are intertwined with cultic and ritual traditions. Language is the medium through which one prays, the vessel for myths and cosmologies. Even in a secular age, language continues to function as a repository of cultural memory, preserving echoes of religious experience long after belief has faded. The simplest example can be seen in the names of days of the week. Although belief in gods such as Saturn and Thor has mostly faded in Europe, their names live on in \textit{Saturday} and \textit{Thursday}, echoing the religious practices of earlier times.

In this modern predicament of ours—what Charles Taylor has famously dubbed a \textit{secular age}—one might be quick to assume that the influence of religion, even in language, belongs entirely to a bygone era. That is, there appears to be a fundamental rupture between a world once organized around religious life—most notably the Church in the European context—and the secular condition that now defines much of contemporary Europe. At first glance, this shift seems to have fundamentally altered the meaning of religious terms. Take, for example, the word \textit{sacrifice}. In contemporary usage, as illustrated by the Cambridge Dictionary example, “She has had to sacrifice a lot for that relationship,” the term refers to personal hardship or emotional commitment rather than to acts of religious devotion \parencite{cambridgeSacrifice}. This secularized meaning contrasts starkly with the original significance of the term. As Benveniste explains in his \textit{Dictionary of Indo-European Concepts and Society}, \textit{sacrifice} once referred to an act performed “to honor the god, to solicit his favor, to recognize his power by means of offerings” \parencite{benveniste2016dictionary}.

This apparent profanization may be more deceptive than it first appears, especially when reconsidering the so-called secular turn. In recent decades, several sociologists and historians have challenged the assumption that modernity entails the disappearance of religion \parencite{terBorg2007nietinstitutionele, white1967, berger1999desecularization}. As Berger remarked, “The world today, with some exceptions to which I will come presently, is as furiously religious as it ever was, and in some places more than ever” \parencite{berger1999desecularization}. If we begin instead with the assumption that religion has not disappeared from modern societies but has rather been reconfigured in new and complex ways, we can return to religious language with a fresh interpretive lens. It is true that figures such as Thor and Saturn are no longer objects of worship, yet the structure of the modern workweek still reflects a Judeo-Christian cosmology in which God created the world in six days and rested on the seventh—a rhythm that continues to shape daily life for many. Furthermore, research on the “fresh-start effect” suggests that people are especially motivated to pursue aspirational goals at temporal landmarks \parencite{freshstart}. The beginning of the week, traditionally associated with the start of creation, becomes a time when individuals seek to recreate or renew themselves by committing to self-improvement goals such as quitting smoking or exercising. Finally, while sacrifice in the past may have been directed toward a deity, it is worth asking to what extent modern forms of sacrifice—to a career, a relationship, or a cause—are as secular as they seem. These commitments often involve orienting one’s life around transcendent values or higher objectives, raising the question of whether they, too, might be understood as religious in character.

To what extent our language continues to evoke explicit or implicit forms of religion across various domains of human activity remains an open and pressing question. As ever more discourse is generated, this question becomes not only suitable for digital methods but increasingly dependent on them. Digital tools allow for the analysis of vast textual corpora, and with the emergence of Large Language Models (LLMs), a new opportunity arises: to explore whether these models, even in a zero-shot setting, can identify language that carries religious significance.

This article represents a first attempt—experimental in nature—at using LLMs alongside a more conventional rule-based approach to detect overt and implicit religious language. Comparing rule-based methods with LLMs opens up avenues to improve our understanding of the conditions under which language is defined as religious. We add to the scarce literature that makes use of LLMs (next to discourse analysis) to analyze text in terms of religious language \parencite{gurlenis2024}. We focus on climate-related discourse used by various NGOs in the Netherlands on their website, which is a meaningful and generative step toward understanding how the sacred might persist in contemporary language.

\subsection{Climate Change and Religion}
Religion and climate change share a complex, historically layered relationship that has long attracted scholarly attention. As early as 1967, White famously argued that Western Christianity’s anthropocentrism and dominion theology significantly contributed to the ecological crisis, and that any lasting remedy must also be religious in nature \parencite{white1967}. Since then, scholars have examined this entanglement from multiple angles: how religious identity shapes views on climate issues, how faith traditions engage with climate theologically, how religion and environmental discourse co-evolve historically and culturally, and even how climate change itself may be interpreted as a kind of religious event \parencite{jenkins2018}.

In examining the relationship between religious language and climate change, we can identify two distinct yet interconnected motifs. The first concerns the application of religious categories—such as \textit{cult}, \textit{prophet}, \textit{Messiah}, \textit{apocalypse}, or even the term \textit{religion} itself—to frame and interpret various dynamics within climate change discourse. These categories are often imposed from outside the discourse, functioning as analytical or rhetorical tools rather than originating from within climate movements themselves. A noteworthy example is the study by \textcite{kyyro2023}, which demonstrates how climate activism is frequently described through conventionally religious terms, casting activists in roles akin to prophets or messianic figures and framing the movement as apocalyptic or cultic in nature. This may be understood as religious language external to climate discourse.

The second motif involves religious language that is autochthonous to climate discourse—that is, language generated organically by stakeholders engaged in climate activism or environmentalist practice. In this case, religious motifs and sensibilities are not imposed from the outside but emerge from within the discourse itself. A compelling example can be found in the interdisciplinary research project \textit{Religious Vocabularies and Apocalyptic Imaginaries of Climate Change in the Netherlands and Beyond}, based at Tilburg University \parencite{vanderStichele2023apocalypse}. Of particular note is Gürlesin’s work, which analyzes the religious dimensions of tweets associated with the Extinction Rebellion movement, providing insight into how religious language and imagination are mobilized by activists themselves.

In his research on the implicit religious underpinnings of online climate change discourse, \textcite{gurlenis2024} draws primarily on the conceptual frameworks of Bailey and ter Borg \textcite{Bailey_2012, terBorg2007nietinstitutionele}. These perspectives provide a compelling lens through which to understand how ostensibly secular movements can adopt narratives and practices that closely resemble those of traditional religions. From Bailey, Gürlesin takes the idea that implicit religion emerges wherever individuals or groups exhibit a deep commitment to ideals, values, or practices that function as sources of ultimate meaning in their lives. Ter Borg’s contribution adds further depth, emphasizing that implicit religion is not limited to shared beliefs or values but also includes the capacity of certain secular phenomena to evoke a “religious promise”—the sense that they offer hope, transcendence, and the possibility of profound transformation.

Gürlesin employs a mixed-methods approach that combines distant and close reading techniques. Much of his computational analysis is facilitated through Atlas.TI, including the use of machine learning models such as XLM-RoBERTa-base for sentiment analysis. By integrating these quantitative methods with qualitative insights drawn from discourse analysis (DA), Gürlesin offers a holistic view of climate-related discourse on X/Twitter. This blended methodology enables a comprehensive exploration of how implicit religious elements intersect with environmental activism on digital platforms.

Regarding his findings, Gürlesin argues that his research offers new insights into the intersection of implicit religion and environmental activism in digital spaces. Through an analysis of XR Nederland’s activity on X/Twitter, he demonstrates that secular movements can take on quasi-religious characteristics, fostering profound emotional and moral commitments among their followers. For instance, phrases such as “We borrow the Earth from our children”  express a moral imperative that transcends individual interests, echoing values commonly found in religious traditions. Similarly, statements like “Nature is a source of inspiration and harmony” attribute a transcendental value to nature, effectively elevating environmental concern to the level of sacred responsibility.

Building on Gürlesin's study, we propose to explore two dimensions that remain unaddressed in his research. First, there is no substantial engagement with large language models (LLMs) as analytical tools for either quantitative or qualitative analysis of the implicit religious underpinnings in online climate change discourse. Given the increasingly ubiquitous role of LLMs in digital research, it is becoming ever more pertinent to experiment with their potential to advance anthropological, sociological, and religious studies. As outlined in the introduction, this article represents a first, exploratory attempt to integrate LLMs—alongside more conventional rule-based methods—for detecting both overt and implicit religious language. Second, Gürlesin’s analysis focuses primarily on tweets related to Extinction Rebellion. While valuable, this focus does not capture the full breadth of climate change discourse. To address this, we expand the scope to include content from a range of NGOs, both secular and explicitly religious, by analyzing their websites, blogs, and articles.

\subsection{Religious language and implicit religion}
We must address the proverbial elephant in the room: what makes language \textit{religious}? Or, posed even more broadly, what is this concept called \textit{religion} that relates to language in such a way that certain uses of it become identifiable as \textit{religious language}? These are not straightforward questions. On the contrary, they form part of a long and contentious intellectual history, marked by deep disagreements over both the definition of religion and the criteria by which language might be considered religious.

\textcite{scott2010} takes religious language to refer to “supernatural agents (God, other deities, angels, etc.), the actions of such agents (miracles, creation, redemption, etc.), and supernatural properties and states of affairs (holiness, heaven and hell, etc.)”. \textcite{vainio2020}, by contrast, argues that references to the supernatural are not a necessary criterion for identifying religious language. He proposes that “religious language consists of sentences that express some claim, belief, attitude, or preference which is religiously relevant”, acknowledging that this definition may seem vague but aligns with the multifaceted nature of religious traditions \parencite{vainio2020}.

Other scholars reject the idea that text content alone determines religious language. \textcite{alston2005} denies that there is such a thing as “religious language,” arguing that no language is used solely for religious purposes. Similarly, \textcite{harrison2007} states that no words are exclusive to religious language, but suggests that something else determines the religiosity of language. For her, this “something else” is principally—though not exclusively—the religious context and/or religious purpose behind a sentence \parencite{harrison2007}. In biblical scholarship, James Barr has famously rejected the assumption that biblical words have a unique or inherently religious meaning. He argued that the biblical language consists of ordinary words from the common vocabulary of ancient Hebrew and Greek and that these words take on religious meaning only when they are used in religious contexts \parencite{Barr1961}.

This discussion becomes even more complex when considered in light of the concept of \textit{implicit religion}. One might feel reasonably confident in identifying statements as religious when they contain explicit references to established religious concepts—for example, “I, as a Muslim, am convinced one seeks God’s help through prayer.” However, the challenge intensifies when we turn to cases of implicit religion, where religious meaning is not overtly expressed in recognizable theological terms. How are we to interpret statements that are not clad in an explicit religious garb, yet seem to evoke a sense of transcendence, moral absolutes, or ultimate concern?

Edward Bailey, a central figure in the study of implicit religion, spent much of his career grappling with the definitional ambiguity of the term. A recurring critique of his approach concerns the perceived risk of overextension: if implicit religion can be found wherever people express deep commitments, integrative foci, and intensive concerns, does everything become religious? Bailey firmly rejected this conclusion. For, the implicit-religion hypothesis merely suggests that any thing may be religious, not the fact that everything is religious \parencite{Bailey_1997}. 

Bailey further acknowledges that some individuals may resent the suggestion that aspects of their behavior contain elements conventionally associated with religion. He addresses this concern in two ways. First, he appeals to the principle of intellectual freedom, arguing that it includes the right of an observer to propose an alternative hypothesis, with the hope that such a claim will open a dialogue between observer and actor \parencite{Bailey_1997}. Second, he situates this practice within a broader historical trajectory, noting that in the last century church historians accepted that secular history and the social sciences could shed new light on the definitions of Christian belief formulated by the Early Church Councils. By analogy, Bailey asks whether, in this century, the study of religion might similarly illuminate aspects of secular life \parencite{Bailey_2012}.

Finally, one might ask whether unanimity or exactitude in defining \textit{religion} is truly necessary for the term to retain meaningful linguistic or analytical relevance. As \textcite{Bailey_2011} aptly observes, “Certainly, the absence of unanimity, or of exactitude, does not prevent its formal and informal use by, say, Departments of Religious Studies.” The utility of the term, in other words, is not diminished by its contested nature; rather, it continues to function productively within both scholarly and public discourse despite its interpretive openness.

Taking our cue from Bailey, in our study we acknowledge the complexities surrounding terms such as \textit{religion}, \textit{religious language}, and related concepts, but we address these challenges in a twofold manner. First, we adopt a more unambiguous approach by working with a curated list of established religious terms, compiled on the basis of expert knowledge. This helps ensure clarity when identifying explicitly religious language. However, we also recognize the limitations of a purely descriptive method. For example, a sentence such as “There lies a cross on the table used for devotional purposes” may contain religious signifiers, but this alone does not necessarily make the language \textit{functionally} religious. Description, in other words, is not always sufficient for interpretation. Second, and more experimentally, we explore whether large language models (LLMs) can serve as tools for identifying both explicit and implicit religious language. In doing so, we do not rely on fixed definitions, nor do we attempt to supply one. Instead, we treat the statistical foundations of LLMs—particularly their capacity to recognize semantic and contextual associations through vector embeddings—as a new kind of hermeneutical aid. LLMs, trained on vast and heterogeneous corpora, can be prompted to identify religious language in a manner akin to consulting a culturally embedded “crowd”: what they return reflects not a single theory, but the accumulated patterns and judgments distributed across countless human texts. This bottom-up approach mirrors how many people already interact with complex concepts like art or religion in everyday life—intuitively, interpretively, and without formal definitions. 

\section{Method}
\label{method}
This study explores how a rule-based model and two different LLMs detect religious language in texts from websites of nine environmental NGOs, including both religious and secular ones. We compiled our data set by scraping website content from the past decade. The rule-based approach to detecting religious language in our data set involves constructing a hierarchical tree of terms derived from literature on religion and ecology. The machine learning approach, on the other hand, uses two different LLMs—GPT-4o mini and Llama 3.3 70B—to identify religious language based on a prompt developed through prompt engineering. The outputs of these models are analyzed through human close reading. This method is described in more detail in the coming paragraphs, followed by the results in Section \ref{sec:results}. Section \ref{discussion} offers a discussion of the findings, including limitations and suggestions for future research, and the article concludes with Section \ref{conclusion}.

\subsection{Data acquisition}
Before collecting data for this study, we first determined which NGOs to include. The included NGOs were selected based on their relevance to environmental causes, the availability of a website in English, and, preferably, a history spanning at least a decade. This is because published material over the last decade would suggest a stable organizational history with consistent activity, proving relevance of the NGO. Furthermore, more data can provide a deeper understanding of the textual content, and opportunities for more meaningful comparisons. We selected NGOs based on the achievement of a balance between religious and secular NGOs and a diversity of religious traditions among religious NGOs. We selected the following secular NGOs: Greenpeace, Extinction Rebellion (XR), World Wildlife Fund (WWF), and Rainforest Alliance. The selected religious NGOs included Christian Climate Action (CCA), A Rocha, GreenFaith, the Indigenous Environmental Network (IEN) and the Interfaith Center for Sustainable Development (ICSD).\footnote{Initially, we also wanted to include the Laudato Si’ Movement in the set of religious NGOs, but we found that their website made use of human verification mechanisms that prevent automated access. As a result, we excluded this NGO from our data set for both ethical and accessibility reasons.} The URLs for each NGO can be found in Table \ref{tab:ngo-stats}

To scrape the content, we used the Internet Archive’s Wayback Machine via its CDX Server API. The CDX Server\footnote{\url{https://github.com/internetarchive/wayback/tree/master/wayback-cdx-server}} is an HTTP-based indexing service used by the Wayback Machine to look up snapshots of web pages. The server returns metadata records in CDX format, including information such as URLs, timestamps, mimetypes, and status codes of archived pages. This enabled us to retrieve a list of all available snapshots of the NGOs’ websites from January 2014 to December 2024 in JSON format, allowing for efficient querying of archived web pages. To retrieve such JSON lists, we sent GET requests\footnote{The following request shows an example of how we retrieved all archived web page URLs from ICSD with mimetype text/html or application/pdf from January 2014 until December 2024, excluding responses with client (4xx) or server (5xx) errors: \url{https://web.archive.org/cdx/search/cdx?url=interfaithsustain.com&matchType=prefix&output=json&from=2014&to=2024&filter=mimetype:(text/html|application/pdf)&filter=!statuscode:^[45]}} to the CDX Server API using the same endpoint structure for each NGO’s website. We used a similar structure for each NGO, accessing all archived webpages\footnote{Once the relevant URLs were identified, we tried to access the archived versions of the websites using the Wayback Machine’s replay URLs:  \url{http://web.archive.org/web/[timestamp]/[original-url]}} belonging to their url (see  Table \ref{tab:ngo-stats}). Due to limitations and access restrictions in the Wayback Machine's infrastructure, we were unable to scrape content directly from these archived pages at scale. As a result, we resorted to scraping the live versions of the URLs instead, using the original URLs retrieved from the CDX Server metadata, rather than the archived pages. Although, we scraped the live versions of the webpages, the CDX metadata remained a useful tool for identifying relevant URLs that had been archived and were likely representative of the NGOs’ key online materials. Thus, this data set consists of a subset of the NGOs’ current web content that is still accessible and archived by the Wayback Machine between 2014 and 2024, offering valuable insight into their recent positions and messaging shared over the past decade. 

This scraped data set consisted of HTML files and PDF files. HTML files were cleaned using the boilerpy3 Python library to retain only the main textual content, with all HTML elements removed. Documents containing only non-English texts were also excluded to ensure that the data set contains mostly English-language material. Detection of non-English texts in the documents was done using the langdetect Python library. However, some documents still contained segments in languages other than English or exhibited residual noise from the scraping process. This final data set was finally split into sentences for further analysis. Table \ref{tab:ngo-stats} presents the size of the data set for each NGO.

\begin{table}[ht]
\centering
\scriptsize
\begin{tabular}{llcc}
\toprule
\textbf{NGO}        & \textbf{Scraped URL}       & \textbf{N documents} & \textbf{N sentences} \\
\midrule
\textbf{Secular NGOs:}    &   &&         \\
\hspace{2em} Greenpeace          & \url{greenpeace.org}             & 9,851   &352,247                      \\
\hspace{2em} XR                  & \url{rebellion.global}           & 2,521   &194,588                      \\
\hspace{2em} WWF                 & \url{panda.org}                  & 773     &18,503                       \\
\hspace{2em} Rainforest Alliance & \url{rainforest-alliance.org}    & 4,624   &243,464                      \\
% \midrule
\textbf{Religious NGOs:}  &      &&               \\
\hspace{2em} CCA                 & \url{christianclimateaction.org} & 452     &10,301                       \\
\hspace{2em} A Rocha             & \url{arocha.org}                 & 342     &6,720                        \\
\hspace{2em} GreenFaith          & \url{greenfaith.org}             & 206     &7,771                        \\
\hspace{2em} IEN                 & \url{ienearth.org}               & 1,345   &49,123                       \\
\hspace{2em} ICSD                & \url{interfaithsustain.com}      & 66      &2,403                        \\
\midrule
\textbf{Total Secular}      &     &\textbf{17,769}              & \textbf{808,802}            \\
\textbf{Total Religious}      & &\textbf{2,411}               & \textbf{77,318}            \\
% \textbf{Total}      &       &\textbf{20,180}              & \textbf{885,120}             \\
\bottomrule
\end{tabular}

\caption{URL link, document counts and sentence counts per NGO.}
\label{tab:ngo-stats}
\end{table}

\subsection{Detecting religious language}
To detect religious language in our data set, we used two different approaches:
(1) a rule-based approach using a hierarchical tree of religious concepts from ecotheology literature and (2) a machine learning approach using two large language models (LLMs). To refine these two approaches, we created a test set by randomly selecting three documents from the total document set of each NGO. These test documents were randomly chosen from the larger data set, and ultimately consisted of 27 documents, totaling 839 sentences from the NGO websites. We decided to detect religious language at the sentence-level as a sentence is more likely to cover one (or few) topic(s) which facilitates our goal. The sentences themselves were not pre-labeled, but were used as raw input to evaluate by close reading whether the approaches meaningfully detected religious language. After close reading, we concluded that the test set contained a diverse range of sentence types, many of which were useful for evaluating our models’ ability to detect religious language.

\subsubsection{Rule-based approach: Hierarchical tree}
\label{sec:rule_based}
\begin{figure}[!htbp]
  \centering
  \includegraphics[width=0.6\textwidth]{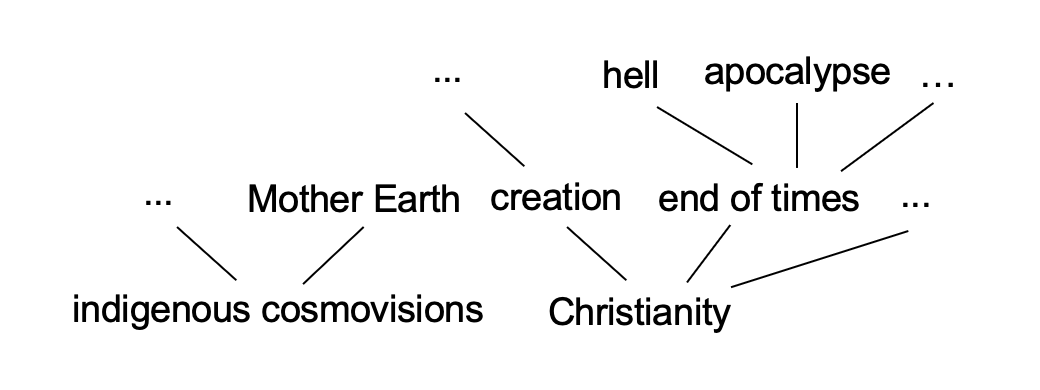}
  \caption{Excerpt from the hierarchical tree with religious concepts.}
  \label{fig:tree}
\end{figure}

For the rule-based approach, we manually constructed a hierarchical tree of religious concepts, organized by religious tradition. This tree is based on two ecotheology handbooks, by \textcite{jenkins2016} and \textcite{conradie2019}.\footnote{For the rule-based approach and the selection of relevant literature we benefited from the useful advice of our colleague Jan Jorrit Haselaar, School of Religion and Theology, Vrije Universiteit.} From \textcite{jenkins2016}, we closely examined the chapters on Christianity, Islam, Judaism, Hinduism, Buddhism, “indigenous cosmovisions,” and “nature spiritualities,” using these as the tree’s roots. Religious concepts from each chapter were assigned to their corresponding root (e.g., “nirvana” under Buddhism), with additional subconcepts added where relevant. We also introduced a “general” root for cross-traditional terms such as “god” and “prayer.” Since the Christianity chapter was more descriptive than conceptual, we supplemented its concepts with those from the handbook on Christian ecotheology by \textcite{conradie2019}. Figure \ref{fig:tree} shows an excerpt from the tree. To improve coverage, we included plural and singular forms of nouns, as well as inflected forms of verbs, where relevant. For example, for the word ``bless", we also included ``blesses", ``blessed", and ``blessing",  all commonly used in religious language.

In preliminary testing, we applied the hierarchical tree to a subset of documents from XR, CCA, and A Rocha, as the full dataset had not yet been collected. We manually evaluated the detected terms and removed those that frequently led to false positives from the tree (e.g., “love,” “hope,” and “submission”). After the full dataset had been collected and the test set created, we re-evaluated the refined tree on the test set and adjusted the tree further based on observed false positives and false negatives.

\subsubsection{Machine learning approach: Large Language Models (LLMs)}
\label{sec:Machine_learning_approach}
The second approach to detect religious language in our data set was a machine learning based approach. In this approach, we let Large Language Models (LLMs) determine for each sentence in the data set whether it contained religious language, along with an argumentation for their answer. As “religious language” is a broad and ambiguous concept, we did not define it beforehand, but initially presented the LLMs with a general prompt:

\begin{tcolorbox}[colback=gray!10, colframe=black, title=General Prompt]
    You are an expert in recognizing religious language within texts. 
Assess whether this text contains any religious language. 
Focus on the overall meaning of the text, not on individual words. 
Be concise. Provide your answer in JSON format. Example:
{"Religious": "Yes or no", "Certainty":"100\%",
"Argumentation": "[Your argumentation here]"}.
Let certainty be a percentage score indicating how certain you are 
of your answer.
\end{tcolorbox}

% \begin{minipage}{\linewidth}
% \begin{verbatim}

% \end{verbatim}
% \end{minipage}

% \begin{lstlisting}

% \end{lstlisting}

We tested the prompt on the subset using three LLMs: GPT-4o mini, Llama 3.2 1B, and Llama 3.2 3B. The Llama models (1B and 3B) were run locally on our machine and GPT-4o mini (hereafter called GPT) was accessed using an OpenAI API key. The Llama models frequently failed to produce coherent reasoning or argumentation, and in some cases, they generated responses based on hallucinated input sentences. We therefore excluded these models from further analysis. GPT, on the other hand, provided coherent answers on the subset sentences and was retained for detailed evaluation.
A close reading of GPT’s outputs revealed two recurring issues. First, the model often labeled purely descriptive references to religious traditions, texts, artifacts, or figures—e.g., ``the apostles and disciples are imprisoned a number of times in the Book of Acts"—as religious language. Second, it failed to recognize terms such as ``sacred earth" or ``Mother Earth" as religious. We argue that this is a significant oversight as, while these terms may not invoke institutional religion, they are rooted in spiritual worldviews, especially those associated with indigenous cosmologies, eco-spirituality, and nature-based belief systems. Similarly, “sacred earth” directly frames the natural world in spiritual or divine terms—an essential characteristic of religious language. Both expressions reflect a reverence for nature that goes beyond metaphor and enters the domain of moral, cosmological, and spiritual meaning. In ignoring these connotations, the model appears to equate religion exclusively with organized, doctrinal traditions (e.g., Christianity, Islam), overlooking non-Western or non-institutional spiritualties. This narrow understanding undercuts the broader conceptual scope of religious language as it appears in environmental discourse.

To refine and optimize our initial prompt, we selected 10 sentences from the test set that illustrated the two issues described above. These 10 sentences are presented in Appendix A under Table \ref{app:A}. During this process, we experimented with various modifications of the prompt and evaluated the responses of the GPT model to the selected sentences. Specifically, we clarified in the prompt that descriptive references to religious traditions, texts, artifacts, or figures should not be classified as religious language unless they conveyed explicit religious meaning or ideas. This definition was added to ensure conceptual clarity and consistency in the model’s interpretation of religious language. Additionally, we incorporated illustrative examples from the test set—both religious and non-religious—to guide the model’s interpretation. These examples were selected to reflect religious expressions and contextual cues, such as doctrinal statements, metaphysical claims, and cultural references. We also experimented with extending the prompt to ask whether a text contained not only religious language but also spiritual language. However, this addition had minimal impact on the responses generated by the LLMs, and we ultimately decided to exclude it from the final prompt.
This process resulted in the following revised prompt:

\begin{tcolorbox}[colback=gray!10, colframe=black, title=Revised Prompt]
You are an expert in recognizing religious language within texts. 
Assess whether this text contains any religious language. 
References to religious traditions, texts, artifacts, or figures, 
when used descriptively, are not to be considered religious language, 
such as in "at least four of St Paul's letters are traditionally called  
the `prison  epistles': Ephesians, Philippians, Colossians, and Philemon" 
or "the Jewish statement was compiled by Professor Nahum Rakover Copied 
with permission from the Alliance of Religions and Conservation (ARC)." 
When the text conveys religious ideas or meaning, it should be 
considered religious language, for example: 
"in `sharing in His suffering' we can share too in His saving mission,"
and "we believe that if the human being accepts this way of life, the  
pain of the planetary wound will be felt, healed, and life  will be born 
again." Provide your answer in JSON format. 
Example: {"Religious": "Yes" or "No", "Certainty": 
"[Percentage]\%", "Argumentation": "[Your argumentation here"}. 
Let certainty be a percentage indicating how certain you are 
of your answer. Be concise in your argumentation.
\end{tcolorbox}

These changes resolved the first issue regarding descriptive references. However, the second issue persisted. Even after rephrasing the task—for example, by replacing “religious language” with “religious and spiritual language” and using alternative example sentences—GPT continued to classify terms like “sacred earth” and “Mother Earth” as non-religious. We therefore proceeded with the version of the prompt given above. 

To enable comparison between the responses of two LLMs, we decided to add another LLM in our further analyses: the Llama 3.3 70B model. As we previously observed, the Llama 3.2 1B and 3B models did not meet the desired performance standards. Therefore, we selected the larger Llama 3.3 70B model (hereafter called Llama) to improve both recognition accuracy and output consistency. This Llama model had a much larger parameter size and hence can better capture complex language patterns and subtle meanings. This allows for more accurate detection of religious language and more consistent adherence to formatting instructions compared to smaller models. Using this Llama model on the test set, we observed that it indeed produced coherent argumentation, contrary to the smaller models. 

\subsubsection{Analysis of the full data set}
Having refined the two approaches to detect religious language, we applied both approaches to the full data set. On sentence-level, we first applied the hierarchical tree to the full data set, collecting all words as found in the tree, the parent paths of the word nodes, and the number of words that were found. Then we applied the machine learning approach on sentence-level by sending requests to OpenAI's Batch API for the gpt-4o-mini model, and requests to Groq's Batch API for the llama-3.3-70b-versatile model. For both these Batch APIs we prepared the batch files with the final prompt as engineered in Section \ref{sec:Machine_learning_approach} as system prompt and the sentences of the NGOs as user prompt.

\section{Results}
\label{sec:results}

\subsection{Overall statistics}
To get an impression of the results obtained using the rule-based and machine learning approaches, several statistics were collected. Table \ref{tab:yesno} shows the percentages of sentences identified as religious and non-religious by the different approaches. Note that the percentages for the LLMs do not sum to 100\%, as some of their responses did not conform to the expected format asked for in the prompt. These responses either did not follow the required JSON structure or failed to provide a clear `yes' or `no' classification. All reported percentages for religious, secular, and total sentences are weighted percentages. 

The percentages indicate that, overall, sentences from religious NGOs are more frequently labeled as religious compared to those from secular NGOs. For IEN, all three classification methods identified a notably lower percentage of religious sentences than for the other religious NGOs. Furthermore, sentences were almost always more frequently labeled religious when using the tree method. Only in the case of CCA did Llama and GPT find more sentences to be religious than the hierarchical tree. For GreenFaith, only Llama found more sentences to contain religious language than the tree. Lastly, the Llama model consistently identified at least as many religious sentences as the GPT model, and in many cases, it labeled a greater number of sentences as religious.

Table \ref{tab:agreement} shows agreement percentages between the three detection methods. In this table, we show the proportion of sentences that were given the exact same label by the different methods. If one of the LLMs failed to return a response in the required JSON format, as specified in the prompt, we classified this as a disagreement between the LLMs. Again, secular, religious and total percentages are weighted percentages. It stands out that agreement for the secular NGOs is much higher than agreement for religious NGOs. The most significant differences in agreement rates between secular and religious NGOs are found in the comparisons between the tree-based method and one of the language models. For all NGOs, agreement between the LLMs is higher than agreement between one of the LLMs and the tree.

Finally, Table \ref{tab:ratio} presents the ratio of `yes' to `no' responses generated by each LLM in cases where the models did not agree (but did give a correct answer, either `yes' or `no'). For example, within the disagreement cases of the XR sentences, this table can be interpreted as: yes occurs 2.74 times as often as no in the responses of the Llama model. These disagreement cases typically reflect more ambiguous or challenging sentences, so this metric offers insight into the tendencies of each model when faced with such cases. The results show that, for nearly all NGOs, Llama was more inclined to classify these sentences as religious, whereas GPT more often classified them as non-religious. The sole exception was WWF, for which GPT gave more religious classifications than Llama.

\begin{table}[]
\centering
\begin{tabular}{l|rr|rr|rr}
\toprule
                    & \multicolumn{2}{c}{\textbf{Tree}} & \multicolumn{2}{c}{\textbf{GPT}} & \multicolumn{2}{c}{\textbf{Llama}} \\
\textbf{}           & \% yes          & \% no           & \% yes          & \% no          & \% yes           & \% no           \\ 
\midrule
\textbf{Secular NGOs:}    &   &&      & & &   \\
\hspace{2em} Greenpeace          & 1.3\%           & 98.7\%          & 0.3\%           & 99.7\%         & 0.4\%            & 99.0\%          \\
\hspace{2em} XR                  & 2.3\%           & 97.7\%          & 0.8\%           & 99.1\%         & 1.2\%            & 98.1\%          \\
\hspace{2em} WWF                 & 2.5\%           & 97.5\%          & 0.2\%           & 99.8\%         & 0.2\%            & 99.5\%          \\
\hspace{2em} Rainforest Alliance & 1.0\%           & 99.0\%          & 0.1\%           & 99.8\%         & 0.1\%            & 99.1\%          \\
\textbf{Religious NGOs:}    &   &&      & & &   \\
\hspace{2em} CCA                 & 23.0\%          & 77.0\%          & 25.4\%          & 74.5\%         & 25.5\%           & 74.0\%          \\
\hspace{2em} A Rocha             & 16.2\%          & 83.8\%          & 13.2\%          & 86.7\%         & 14.9\%           & 84.9\%          \\
\hspace{2em} GreenFaith          & 20.5\%          & 79.5\%          & 19.7\%          & 80.0\%         & 23.6\%           & 76.0\%          \\
\hspace{2em} IEN                 & 6.0\%           & 94.0\%          & 2.7\%           & 97.1\%         & 2.9\%            & 96.8\%          \\
\hspace{2em} ICSD                & 36.2\%          & 63.8\%          & 27.1\%          & 72.8\%         & 32.6\%           & 67.1\%          \\
\midrule
\textbf{Total Secular}    & 1.5\%           & 98.5\%          & 0.4\%           & 99.6\%         & 0.5\%            & 98.8\%          \\
\textbf{Total Religious}  & 11.9\%          & 88.1\%          & 9.4\%           & 90.4\%         & 10.4\%           & 89.3\%          \\
\textbf{Total}      & 2.4\%           & 97.6\%          & 1.2\%           & 98.8\%         & 1.4\%            & 98.0\%         \\
\bottomrule
\end{tabular}
\caption{Percentages of sentences recognized as religious language (yes) or not (no) by the different models per NGO.}
\label{tab:yesno}
\end{table}

\begin{table}[ht]
\centering
\begin{tabular}{l|c|c|c|c}
\toprule
                    & \textbf{Overall} & \textbf{Tree \& } & \textbf{Tree \& } & \textbf{GPT \& } \\ 
                    & & \textbf{GPT} & \textbf{Llama} & \textbf{Llama} \\ 
\midrule
\textbf{Secular NGOs:}    &   &&      &  \\
\hspace{2em} Greenpeace          & 97.9\%           & 98.7\%               & 98.0\%                 & 99.0\%                \\
\hspace{2em} XR                  & 96.7\%           & 97.9\%               & 97.0\%                 & 98.4\%                \\
\hspace{2em} WWF                 & 97.2\%           & 97.5\%               & 97.3\%                 & 99.5\%                \\
\hspace{2em} Rainforest Alliance & 98.1\%           & 98.9\%               & 98.2\%                 & 99.1\%                \\
\textbf{Religious NGOs:}    &   &&      &   \\
\hspace{2em} CCA                 & 81.2\%           & 86.3\%               & 86.0\%                 & 90.1\%                \\
\hspace{2em} A Rocha             & 90.9\%           & 93.0\%               & 93.2\%                 & 95.4\%                \\
\hspace{2em} GreenFaith          & 78.0\%           & 84.4\%               & 81.1\%                 & 90.3\%                \\
\hspace{2em} IEN                 & 94.5\%           & 95.7\%               & 95.2\%                 & 98.0\%                \\
\hspace{2em} ICSD                & 70.1\%           & 77.4\%               & 75.7\%                 & 87.0\%                \\
\midrule
\textbf{Secular Total}    & 97.6\%           & 98.5\%               & 97.8\%                 & 98.9\%                \\
\textbf{Religious Total}  & 89.6\%           & 92.3\%               & 91.4\%                 & 95.5\%                \\
\textbf{Total}      & 96.9\%           & 98.0\%               & 97.3\%                 & 98.6\%      \\         
\bottomrule
\end{tabular}
\caption{Percentages of agreement between the different models per NGO.}
\label{tab:agreement}
\end{table}

\begin{table}
\centering
\begin{tabular}{lcc}
     
\toprule
\textbf{NGO}        & \textbf{GPT} & \textbf{Llama} \\
\midrule
\textbf{Secular NGOs:}    &   &       \\
\hspace{2em} Greenpeace          & 0.38         & 2.62           \\
\hspace{2em} XR                  & 0.36         & 2.74           \\
\hspace{2em} WWF                 & 1.15         & 0.87           \\
\hspace{2em} Rainforest Alliance & 0.80         & 1.24           \\
\textbf{Religious NGOs:}    &   &       \\
\hspace{2em} CCA                 & 0.98         & 1.02           \\
\hspace{2em} A Rocha             & 0.44         & 2.26           \\
\hspace{2em} GreenFaith          & 0.41         & 2.46           \\
\hspace{2em} IEN                 & 0.76         & 1.32           \\
\hspace{2em} ICSD                & 0.38         & 2.59           \\
\midrule
\textbf{Secular Total}    & 0.41         & 2.41           \\
\textbf{Religious Total}  & 0.62         & 1.6            \\
\textbf{Total}      & 0.51         & 1.93           \\
\bottomrule
\end{tabular}
\caption{   Ratio of yes/no responses in disagreement cases: ``yes occurs $x$ times as often as no in the responses of $y$ language model.''}
\label{tab:ratio}
\end{table}

\subsection{Disagreements between the tree and the LLMs}
\label{disagreement_tree_LLMs}

As the number of sentences in the results was vast, we focused our qualitative analysis primarily on cases where the detection methods disagreed. Such cases are particularly informative because they highlight where and how the three models diverge in their interpretation of religious language, potentially revealing the strengths, assumptions, and limitations of each method.

We begin by discussing the insights gained from analyzing disagreements between the tree and both LLMs. Close reading of these cases revealed many instances where the tree detects a religious term, but the LLMs do not label the sentence as religious, arguing that it contains only descriptive references to religious traditions or concepts. For example, the words “Christian,” “Jewish,” “Muslim,” “Hindu,” and “Buddhist” are often viewed by the LLMs as used descriptively, for instance in references to “Christian organizations” or “Muslim countries.” Other terms frequently seen as descriptive by the LLMs include the names of sacred books or the word “ritual,” such as in:

\begin{verbatim}
"Other memories come back to me: a welcome in Birmingham 
Progressive Synagogue as they built their Sukkah; a local
mosque with a reading from the Koran about the environment." 
(CCA)
\end{verbatim}

or
\begin{verbatim}
"The net is used for Bluefin tuna in a Mattanza fishery 
(ancient fishing ritual).(WWF)
\end{verbatim}

While such sentences are not labeled as religious by the LLMs, the terms detected by the tree are often acknowledged in the LLMs’ responses. However, the models argue that these terms are merely used descriptively, which agrees with our prompting efforts to rule out purely non-religious descriptive usage of religious terms such as titles or names.

In some cases, the terms identified by the tree are not mentioned at all by the LLMs. Examples of words where this happens often are “sacrifice” and “devote,” and their derivatives, such as in:

\begin{verbatim}
"75% of agriculture land is devoted to raising animals."
(Greenpeace)
\end{verbatim}
or
\begin{verbatim}
"This dirty deal sacrifices communities of color that are 
already carrying the burden from decades of environmental 
injustice while driving all of us even further into climate 
devastation.(IEN)
\end{verbatim}
In these cases, the LLMs typically state that the sentence does not contain any religious ideas or meanings, without referencing the potentially religious connotations of the words.

An additional interesting category of disagreement is quotations from sacred books. If these lines do not contain any of the tree’s predefined keywords, the tree does not mark them as religious, but there are instances where the LLMs do recognize them. An example is a line from Psalm 121:

\begin{verbatim}
"I look at the hills: where does my help come from?(CCA)
\end{verbatim}

Both GPT and Llama label this sentence as religious, but with different reasoning. Both GPT and Llama explicitly identify it as Psalm 121:1, while GPT additionally explains why it aligns with the definition of religious language given in the prompt:

\begin{verbatim}
"The phrase `where does my help come from?' is reminiscent of
biblical language, specifically from the Psalms in the Old 
Testament, which often reflects a search for divine assistance 
or guidance.(GPT)
\end{verbatim}

\subsection{Disagreements between the LLMs}
Next, we discuss the insights gained from analyzing cases of disagreement between the two LLMs. Important to note is that these example cases are instances where the language models did not agree in their final response. Therefore, the findings based on these differences can not be taken as representative or generalizable to the models’ overall behavior, but solely as difference in behavior within disagreement cases. 

\subsubsection{Incorrectly formatted responses}
An agreement rate that stands out among the secular NGOs in Table \ref{tab:agreement} is the slightly lower agreement in the XR sentence inputs. However, inspection of the LLMs's responses revealed that for many of the disagreement cases, Llama generates an output that is not consistent with the JSON format.  With close reading it was seen that erroneous outputs are often found for very short input sentences, sentences that are difficult to interpret without any additional context, or input sentences that may be interpreted to be a prompt themselves by the LLMs, such as in:
\begin{verbatim}
"Sure.(XR)
\end{verbatim}
and
\begin{verbatim}
"What else can you do?(CCA)
\end{verbatim}

The percentage of Llama's incorrectly formatted responses is particularly high for XR. For this NGO, 1,467 times out of all 3,138 cases Llama does not respond in the requested format, which is almost half of the disagreement cases. GPT only gives incorrectly formatted responses in 174 times for these XR sentences. Out of a total of 21,310 disagreement cases, Llama generated 5,660 responses in an incorrect format (26.6\%) and GPT 704 (3,3\%).

\subsubsection{Types of argumentation}
Closely reading the LLMs' responses revealed different types of argumentation, but these types do not seem to be strongly linked to either one of the LLMs. That is, within the disagreement subset, GPT and Llama use similar reasoning when explaining their classification of a sentence as religious or not. Moreover, recognition of specific words or phrases does not seem to be linked to the model that recognized it, despite Llama more often recognizing the sentences to be religious. An example can be found in argumentation for the following sentence:

\begin{verbatim}
"Goddamn weather channel, never right.(XR)
\end{verbatim}

This short phrase occurs 7 times in the disagreement cases, all originating from the XR data. Out of 7 times, Llama labels this phrase as religious 4 times, while GPT labels it religious the other 3 times. The LLMs give similar arguments on why this sentence contains religious language or not. When arguing the sentence does contain religious language, GPT states that this sentence ``invokes the name of God, which is a religious reference", hence it ``reflects a use of religious language". In cases where Llama classifies this sentence as religious, it gives a very similar argumentation, also saying it ``invokes the name of God", thus being ``an example of religious language". In argumentation against this sentence containing religious language, GPT argues that the use of God is ``used in a secular context", and is ``used as a curse rather than a reference to religious beliefs or practices". Llama similarly states that ``the term `Goddamn' is used as an expletive", and does not ``convey religious ideas or meaning". 

Overall, with close-reading, we observed that several reoccurring types of argumentation seem to be used by both LLMs.

\paragraph{Metaphorical interpretation}
A frequently recurring argument in both LLMs, when they label a sentence as non-religious while the other labels it as religious, is that what the other model interprets as a religious expression is interpreted by this model as metaphorical. This type of argumentation is not unique to either LLM, as illustrated by the examples below.

\begin{verbatim}
"And still, on the altar of our greed the children’s lives are 
laid.(CCA)
\end{verbatim}
In this disagreement case, Llama argues that ``the phrase `altar of our greed' is a metaphorical expression”, whereas GPT finds the phrase evokes ``religious imagery”. However, GPT uses a similar reasoning as Llama does, for the following sentence:

\begin{verbatim}
"This makes it easy to dismiss the experience of others who see 
the ‘groans of creation’ and how it impacts the most
marginalised.(CCA)
\end{verbatim}
GPT argues that ``the phrase `groans of creation' could be interpreted as a metaphorical expression related to suffering and ecological concerns, but it does not explicitly convey religious ideas or meanings.” In contrast to GPT, Llama finds “groans of creation” to be “a biblical reference to Romans 8:22, implying a religious connotation.”\footnote{Cf. Romans 8:22 "We know that the whole creation has been groaning together as it suffers together the pains of labor" (NRSV)}

Thus, in these cases, the different models rely on metaphorical interpretation as a basis for their classification decisions, but come to different conclusions, highlighting that this form of argumentation is not model-specific but shared across models. However, even if shared across models, the way how thay apply this interpretation varies between the models.

\paragraph{Sacred texts}
In this previous example, only Llama recognizes “groans of creation” as a biblical reference to Romans 8:22. This pattern seems to happen more often, where GPT does not mention a biblical verse or passage, whereas Llama does. As noted above, in Section \ref{disagreement_tree_LLMs}, Llama also identified Ps 121:1 as a biblical quotation. This is also the case for:

\begin{verbatim}
Yet our knee of oppression remains on their neck as they cry to 
us for help.(CCA)
\end{verbatim}
and
\begin{verbatim}
"All th e Ear th Bel on gs t(CCA) 
\end{verbatim}
For both cases, Llama mentions a reference to a specific biblical passage; for the first example Psalm 107:6\footnote{"Then they cried to the L{\textsc{ord}} in their trouble, and he delivered them from their distress." (NRSV)} or Exodus 2:23\footnote{"The Israelites groaned under their slavery and cried out. Their cry for help rose up to God from their slavery." (NRSV)} and for the second example Psalm 24:1\footnote{"The earth is the L\textsc{ord}’s and all that is in it, the world, and those who live in it." (NRSV)}. GPT does not mention this biblical passage, arguing that both sentences do not ``explicitly convey religious ideas or themes". Llama does mention that the phrase ``cry to us for help" has a biblical connotation, but ``without more context, it's unclear if the author intends a direct religious reference or is using the phrase metaphorically". Interestingly, despite the mention of limited context, Llama does label the sentence to be religious, seemingly due to the reference to the biblical passage. 

For the second example GPT, on the one hand, recognized the phrase ``all the Earth Belongs to", and mentions it does ``not convey any specific religious ideas or meanings". It mentions that is more likely to reference ``a general concept of belonging or stewardship", but denies an expression of religious beliefs or sentiment. Llama, on the other hand, recognizes the phrase ``all the Earth Belongs", which it mentions to be a quote from Psalm 24:1, thus ``suggesting a religious connotation".

In the dataset, there are instances where both LLMs mention a specific Bible reference, but we found no instances in the disagreement data set where GPT mentions a biblical reference while Llama does not. However, due to the large size of the data set, this observation is difficult to confirm with certainty.

Other interesting examples where the LLMs refer to sacred books in their argumentation can be found in the following sentences. 
\begin{verbatim}
That ought to have raised serious concerns for Genesis.(Greenpeace)
\end{verbatim}
GPT's response to this sentence is that Genesis is a reference to a biblical book, ``which is a foundational text in religious traditions, particularly in Judaism and Christianity''. Hence, it concludes that the context of the sentence ``implies a significance linked to religious narratives". Llama, on the other hand, responds that Genesis is only used ``descriptively without conveying religious ideas or meaning".

An inverse example of argumentation can be found for the following sentence.
\begin{verbatim}
Part of that is language restoration and part of that is 
understanding that genesis of who we are.(IEN)
\end{verbatim}
GPT argues that the text does not ``convey religious ideas or meanings, even though it references `genesis,' which can have religious connotations''. Llama argues that the term `genesis' has a strong biblical connotation, and that even though ``the religious connection is implicit'', the use of the term genesis still invokes a religious meaning. 

\paragraph{Descriptive use}
Another common argument used by both LLMs is that a sentence does not contain religious language because it is purely descriptive. An example of this type of reasoning can be found for the following sentence:
\begin{verbatim}
"You can send messages of support to Rev Bill at 
billwinprison@protonmail.com Find out more about Just Stop Oil 
via a Zoom talk: Our Responsibilities at This Time on Thursday 
26th May at 7.00pm.(CCA)
\end{verbatim}
GPT classifies this sentence as religious since the mention of ``Rev Bill" indicates ``a reference to a religious title (Reverend) commonly used within Christian traditions, which suggests a religious context". Llama, however, considers these ``references to a person with a religious title (Rev Bill)'' merely descriptive, not conveying ``any religious ideas or meaning".

The following sentence forms an inverse example: 
\begin{verbatim}
He is Chaplain to Gypsies and Travellers in Salisbury Diocese 
and lives at Hilfield Franciscan Friary.(CCA)
\end{verbatim}

Here, GPT argues that the text does not contain religious language. It acknowledges that it contains ``references to a chaplaincy and a friary, which are associated with religious roles and communities", but argues that it merely ``describes a person's role and living situation," thus not conveying religious ideas or meaning. This response is notable as in the first example, a mere title is sufficient for GPT to detect a religious context. These examples highlight how GPT applies its classification criteria inconsistently when analyzing and interpreting religious references embedded in seemingly descriptive statements.

For this second example sentence, Llama argues that the text mentions a chaplain and a Franciscan friary, which are ``explicitly religious terms, indicating the presence of religious language". This argumentation also contrasts its response to the first example, which illustrates an inconsistency in its application of classification criteria.

\subsection{Notable cases: Mother Earth and Sacred Earth}
As indicated above (see Section \ref{sec:Machine_learning_approach}), terms such as the “sacredness of the earth” and “Mother Earth” were not recognized as religious by GPT in the test set. However, they play a role in various religious traditions. Admittedly, it is debated to what extent the diverse mother figures in myths from all around the world can been taken together as representing an almost universal Mother Earth figure. It has been argued that the latter is rather a Western intellectual invention, that only in the 1970s came into usage among North American and Australian indigenous religions, “a scholarly invention in service to the comparative enterprises of essentialist, patternist, and encyclopedic studies scantly supported by reliable accurate descriptions of cultural and historical reality" \parencite{Gill2024}, see also \parencite{Gill1987, Gill1990}. But even if we would admit that Mother Earth has been invented in the “creative encounter among native Americans, European-descended colonialists and (post-colonial) environmentalists, who have conspired in the conception of mother earth” \parencite{Sundström_2024}, in their current usage, these concepts have clear religious overtones. They are used in modern Mother Earth spirituality \parencite{sjoo1987great} and have been incorporated in contextual African Bible interpretation against the backdrop of environmental challenges \parencite{Leshoto2021}, alongside with other indigenous religious concepts such as Ubuntu \parencite{taringa2025mother}. Could it be that the LLMs identify “religion” mainly as belonging to one of the five so-called word religions (Judaism, Christianity, Islam, Hinduism, Buddhism)? \footnote{Note that the paradigm of world religions itself is claimed to have been shaped by European universalist biases \parencite{segal2007invention}.} 

Because “Mother Earth” and “sacred earth” stood out during prompt engineering as terms GPT did not consider religious, we analyzed all sentences containing either term in the full data set to examine how the different detection methods responded.

\subsubsection{Sentences mentioning “Mother Earth”}
The full data set contains 1,229 sentences that mention “Mother Earth,” drawn from the websites of Greenpeace, GreenFaith, XR, and IEN. As the term is included in the hierarchical tree, all of these sentences are automatically labeled as religious by the rule-based method. By contrast, GPT labels only 415 (33.8\%) of them as religious, and Llama 457 (37.2\%), with 326 cases where they both label as religious.

When the LLMs classify these sentences as non-religious, they often acknowledge that the term “Mother Earth” may carry “spiritual connotations” or similar associations, but argue that it does not explicitly express religious ideas or meaning. However, in other cases, the LLMs do conclude that the sentence conveys religious meaning. An illustrative case is a sentence that appears 12 times in the data set, as it is included on multiple webpages under different URLs:

\begin{verbatim}
My heart beats rapidly in my chest, and I feel like I can 
faintly hear Mother Earth’s heart beating rapidly far below me. 
(XR)
\end{verbatim}
Both LLMs are inconsistent in labeling this sentence. Of the 12 instances, GPT labeled it as religious twice and Llama seven times. We show two types of argumentation GPT gave—one for labeling the sentence as religious and one for labeling it as non-religious—to illustrate this inconsistency. Llama provided similar reasoning.

When labeling the sentence as containing religious language, GPT explained:
\begin{verbatim}
The phrase `Mother Earth’ refers to a spiritual concept of the 
Earth as a nurturing figure, which can convey religious themes 
of connection to nature and reverence for the planet. This 
suggests a relationship that transcends mere description, 
implying a spiritual or religious connection.(GPT)
\end{verbatim}
When labeling the same sentence as non-religious, GPT gave the following argumentation: 
\begin{verbatim}
The mention of `Mother Earth' suggests a connection to nature 
and possibly animistic or eco-spiritual ideas; however, it does 
not convey explicit religious ideas or meaning associated with 
specific religious traditions. It is more poetic than doctrinal.
(GPT)
\end{verbatim}

In addition to revealing many inconsistencies in the labeling of the LLMs, close reading of the LLMs' arguments revealed two factors that, when the LLMs do label a sentence mentioning ``Mother Earth" as religious, are often mentioned in their reasoning. These factors are (1) a (strong) personification of “Mother Earth” and (2) the presence of a second religious element. An example of the first factor can be found in Llama's response to the following sentence:
\begin{verbatim}
A scar cutting across Mother Earth, drawing the blood from
within her to the surface.(XR)
\end{verbatim}
Llama specifically mentions the personification in the argumentation for labeling the sentence as religious:
\begin{verbatim}
The text uses a metaphor that personifies the earth as `Mother 
Earth', implying a spiritual or sacred connection, which is a 
common concept in some indigenous and pagan religious traditions.
(Llama)
\end{verbatim}
An example of the second factor—an additional religious cue—can be seen in the following sentence, where GPT picks up on the speaker's title:
\begin{verbatim}
``We, the people, need to live and now start to protect mother 
earth,'' Rev Dr Neddy Astudillo said.(GreenFaith)
\end{verbatim}
GPT gives the following explanation for classifying the sentence as containing religious language: 
\begin{verbatim}
The use of `Rev' indicates a religious title, suggesting that
the speaker holds a position within a religious context. 
Additionally, the phrase `mother earth' often carries spiritual 
connotations, especially within various religious and indigenous 
belief systems,thus conveying a broader ethical imperative that 
is rooted in religious or spiritual thought.(Llama)
\end{verbatim}
However, it should be noted that these two factors by no means account for all differences in how sentences containing “Mother Earth” are labeled. As the example sentence occurring 12 times aptly illustrates, the LLMs display considerable inconsistencies in their labeling.

\subsubsection{Sentences mentioning “sacred earth”}
The full data set contains 52 sentences that include the phrase “sacred earth,” drawn from the websites of GreenFaith, ICSD, and IEN. Because the term “sacred” appears in the hierarchical tree, all of these sentences are labeled as religious by the rule-based method. The LLMs behave differently, however, with GPT labeling only 6 (11.5\%) of them as religious and Llama labeling 39 (75\%) as religious.

When GPT labels these sentences as non-religious, the term “sacred” is rarely mentioned at all in its argumentation, with the model simply stating that the sentence does not convey religious ideas or meaning (or using similar phrasing). In the six instances where GPT does label the sentence as religious, another religious element is present in the sentence, such as a reference to the Quakers or to a ritual, and this is addressed in the argumentation.

When Llama labels sentences containing ``sacred earth" as religious, the argumentation often implies that the use of the word ``sacred" is enough. However, Llama is not consistent in this, as the following sentence in the data set highlights:
\begin{verbatim}
For Our Sacred Earth
Aug 28, 2024
Climate change, driven by fossil fuels, threatens the well-being 
of people and planet alike.(GreenFaith)
\end{verbatim}
This sentence occurs 39 times across different URLs and is always labeled as non-religious by GPT. However, Llama labels 35 instances as religious, with argumentation such as 
\begin{verbatim}
The use of the word `Sacred' implies a reverence or spiritual 
significance, suggesting a religious connotation.(Llama)
\end{verbatim}
However, Llama classifies the exact same sentence as non-religious four times, arguing that, though the term ``sacred" can have religious connotations, ``it appears to be used descriptively to convey a sense of importance and value" without conveying religious ideas or meaning. 

The ambiguity regarding "Mother Earth" and "sacred earth" can also be observed with another term originating from indigenous cultures, namely "Ubuntu". In scholarly literature Ubuntu has been labelled an African "ethics", "philosophy" or "worldview", among others. It has been integrated in Christian religious thinking in the "Ubuntu Theology" developed by Desmond Tutu. It occurs once in the dataset in the sentence

\begin{verbatim}
The alternative model for Africa’s development can be shaped 
by the spirit of ubuntu, driven by ubunifu and bound together by
ingwebuike.(GP)
\end{verbatim}

Llama labels this sentence as non-religious and gives the following explanation:

\begin{verbatim}
The terms `ubuntu', `ubunifu', and `ingwebuike' are cultural 
concepts originating from African traditions, but in this 
context, they are used descriptively to describe a development
model, without conveying any specific religious ideas or 
meanings.
\end{verbatim}

GPT, however, labels the sentence as religious because Ubuntu "conveys a spiritual and ethical idea that is deeply ingrained in many African cultures."

\section{Discussion}
\label{discussion}

\subsection{Context}
The results show that the hierarchical tree labels sentences as containing religious language more frequently than either LLM across the entire dataset. This might be counterintuitive, given that the tree's vocabulary is limited and cannot account for all possible indicators of religious language. Still, the tree classifies more sentences as religious language than the LLMs. A possible explanation can be found in the fact that LLMs consider the broader sentence-level context, not just individual words. As a result, they often interpret instances where the tree identifies ``religious terms" as merely descriptive or not used in a religious sense. However, the LLMs did require the prompt to specify that descriptive use should not count, suggesting that they initially considered those instances as religious language, until told otherwise.

The LLMs can, however, only take limited context into account, as they are only fed one sentence at a time. The broader text or the context in which the sentence was originally uttered cannot be considered, at least not beyond what can be deduced from the sentence itself. The LLMs do occasionally indicate in their argumentation that they would need such broader context in order to determine whether a sentence contains religious language. However, in most cases, they rely on the presence of particular words or combinations of words within the sentence to make this determination. This seems to align more closely with views that treat religious language as determined by textual content itself \parencite{scott2010, vainio2020}, rather than with views that argue for a more contextual dimension to what makes language religious \parencite{alston2005, harrison2007, Barr1961}; cf. Section \ref{introduction}.

\subsection{GPT vs. Llama}
Furthermore, comparing the two LLMs, we find that Llama classifies sentences as religious more frequently than GPT—both across the entire dataset and within the subset of disagreement cases. Both models rely on a range of argumentation strategies, including dismissing ``religious terms" as descriptive or metaphorical, but no specific strategy appears unique to one model. This makes it difficult to determine why Llama is more inclined to label text as religious.

One notable difference between the two LLMs concerns their detection of biblical texts. Although this applies to only a small subset of the dataset and therefore cannot fully explain why Llama labels sentences as religious more often than GPT, it is nevertheless a revealing pattern. We found several instances in which Llama explicitly identifies a reference to a specific biblical passage, while GPT does not. While there are cases where GPT does mention a specific scripture, we found no examples where GPT identified a biblical reference that Llama did not also detect. This suggests that Llama may be more sensitive to detecting biblical references or allusions, raising the question of what accounts for this difference. Possible explanations could lie in differences in training data, model architecture, or fine-tuning, but exploring this lies beyond the scope of the present study.

\subsubsection{Inconsistencies in the LLMs' argumentation}
The labeling and argumentation of both LLMs reveal notable inconsistencies, most clearly visible in their responses to duplicate sentences within the dataset. Inconsistencies are also evident in the LLMs' differing responses to sentences containing the terms ``Mother Earth" or ``sacred earth". During prompt engineering on the test set, GPT did not classify such sentences as religious, yet in the full dataset, both LLMs occasionally do.

Specifically, GPT labels 33.8\% of sentences mentioning ``Mother Earth" as religious, while Llama labels 37.2\%. In their justifications, both models acknowledge the term’s spiritual connotations, but this does not consistently lead to a religious classification. The surrounding sentence context seems to play a role, yet it cannot fully account for the variation, given that even identical sentences are inconsistently labeled. It is at least not immediate for either LLM that ``Mother Earth" signifies religious language, with most such sentences labeled as non-religious. A clearer difference appears in responses to “sacred earth”: GPT classifies 11.5\% of these sentences as religious, while Llama does so in 75\% of cases. GPT often overlooks the term “sacred” entirely in its justifications, whereas Llama appears more inclined to treat “sacred” as a sufficient reason to label the sentence as religious—though still inconsistently.

While these inconsistencies make strong conclusions difficult, they are themselves revealing. It might be said that the LLMs reflect the difficulty of determining what religious language is, especially when it comes to terms more associated with spirituality than with traditional religion. Based on the terms examined here, Llama appears more inclined to treat such language as religious, though further research would be needed to generalize this observation.

\subsection{Zero-shot detection of implicit religion}

The inconsistencies outlined above—in how LLMs classify terms like “Mother Earth” or “sacred earth,” and in their responses to duplicate or near-identical sentences—highlight a deeper issue: large language models appear to lack a stable, nuanced framework for identifying implicit religious language. This observation becomes especially significant when we consider the models' behavior in a zero-shot setting. Operating without prior task-specific training or fine-tuning, the models respond based solely on the statistical associations they have acquired from large-scale, general-purpose corpora.

This zero-shot context is revealing. Precisely because no additional domain-specific instructions are provided, the models' classifications reflect broader cultural-linguistic patterns and biases embedded in their training data. While terms like “Mother Earth” or “sacred” may carry religious or quasi-religious connotations, especially in environmentalist discourse, LLMs often hesitate to label them as such. The models tend to privilege explicit, conventional religious markers—such as references to God, scripture, or institutional affiliations—while overlooking more subtle forms of religious expression, such as spiritualized language, symbolic metaphors, or references to transcendental values.

In this sense, the models’ default behavior reveals a divergence from interpretive frameworks like Bailey’s theory of implicit religion, which emphasizes how religious meaning can emerge in secular contexts through deeply held commitments, rituals, or orientations toward ultimate meaning. Whereas Bailey positions explicit religion as merely one expression of a broader implicit religiosity, our findings suggest that LLMs often reverse this logic \parencite{Bailey_1997, Bailey_2011}. For the models, religious tends to mean explicitly religious—a stance that aligns more closely with conventional usage and less with postsecular accounts of religion.

Leading into the next section, this finding raises a broader epistemological concern: if LLMs reflect prevailing textual norms, then it is worth asking whether our general cultural and linguistic conception of religion—as encoded in vast digital corpora—still systematically excludes or marginalizes implicit forms. The unmodified behavior of the model, then, may not merely reflect a technical limitation, but could also offer a suggestive glimpse into how religion is popularly and statistically constructed in language.

\subsection{Limitations and suggestions for further research}
\label{sec:limitations}
Several limitations of this study should be acknowledged. Firstly, the LLMs were used in a zero-shot setting—that is, they were given a specific prompt without any task-specific fine-tuning. While this approach demonstrates the generalizability and adaptability of LLMs, it also highlights their sensitivity to prompt design. We do not claim that the prompt used in this study is the only or optimal one for detecting religious language; different phrasings could lead to substantially different outputs. Secondly, there remains an important distinction between what LLMs say in their explanations and how they actually arrive at their classifications. The reasoning they provide may not accurately reflect the internal processes or statistical associations used to generate their labels. Finally, the qualitative observations presented in this paper are based on close readings of a subset of the results. A more comprehensive analysis of the full dataset may reveal additional patterns or nuances not captured here.

These limitations also point toward promising directions for future research. One avenue would be to explore what is known about the internal workings and training data of each LLM, in order to reinterpret their classifications and reasoning in that light. %A specific suggestion is that we specifically ask the LLMs to give a certainty percentage of their answer in the prompt. Within this study we ultimately did not use this metric, however it can still bring light to the inner workings of the LLM, how the LLMs reasons, specifically in combination with their argumentation, and how it interprets religious language.
Another direction would be to examine more thoroughly how LLMs respond to language associated with spirituality or animism in contrast to language tied to so-called “world religions”, to assess whether particular types of religious or spiritual language are more easily recognized. 

\section{Conclusion}
\label{conclusion}
This study explored how LLMs and a rule-based method, based on literature, identify religious language within climate discourse. We investigated to what extent LLMs can meaningfully recognize religious language, and how their interpretations reflect assumptions about what religious language is. We specifically studied the domain of climate discourse, as its complex relationship with religion offered a suitable testing ground to investigate the intricacies of religious language detection. Specifically, we scraped textual content from websites of nine environmental NGOs, both religious and secular, from the past decade. At the sentence level, we analyzed how NGOs, both religious and secular, use explicit and implicit religious language to convey their standpoints on climate, and how this language is interpreted by both a rule-based method and LLMs.

Overall, we found that the rule-based approach, despite being based on a fixed vocabulary, labeled more sentences as religious than the LLMs. The LLMs showed inconsistencies in classifying similar or even identical sentences, and required prompt clarification to avoid wrongly labeling descriptive references as religious language. These findings show the difficulties in detecting religious language, even when using LLMs with contextual reasoning. The results reflect a broader tension in debates on religious language, namely, whether religious language can be identified purely through vocabulary, or whether context and intent of the language must be considered. The LLMs mostly labeled sentences as religious based on explicit textual features, such as religious vocabulary or references to sacred texts. Sometimes, however, the LLMs noted that they needed more context to determine whether the sentence contained religious language. 

Furthermore, this study shows the possibilities and limitations of using LLMs as tools for detection of religious language. On the one hand, LLMs can be a useful tool to explore the use of religious language in climate discourse. On the other hand, the inconsistencies in the argumentation of the models and their sensitivity to the used prompt show the risks of using LLMs in the detection of such complex language. Overall, we can conclude that the LLMs alone cannot be used as objective classifiers. For now, they can only be used as tools that reflect and reproduce the ambiguities and biases present in their training data and the given prompt. The responses these LLMs generate reflect the intricacies of detecting and defining religious language, within complex contexts such as climate discourse.

\section*{Acknowledgements}
The authors thank the Vrije Universiteit Amsterdam for providing financial resources for the study.

\section*{Data availability}
All data used in this paper will be made publicly available via OSF at the time of publication of the study.

\section*{Code Availability}
All codes and scripts developed and used for statistical analysis will be made publicly available via OSF at the time of publication of the study.

\section*{Competing Interests}
The authors do not have any competing interests to declare.

\newpage
% \begin{landscape}
% \appendix
% % \label{app:A}
% \section{Ten test sentences used for prompt engineering}
% \begin{appendices} 
\section*{Appendix A: Ten test sentences used for prompt engineering}
\setcounter{table}{0}
% \addcontentsline{toc}{section}{Appendix A - Ten test sentences used for prompt engineering}

% \usepackage[normalem]{ulem}
% \usepackage{tabularray}
% \begin{table}[ht]
%     \centering
%     \tiny
    
    % \begin{longtable}{p{8cm}p{2cm}p{5}}
    % \begin{table}[h]
% \scriptsize
% \centering
% \caption{The ten test sentences selected from the test on which the prompt was engineered.}
% \begin{tblr}{
%   width = \linewidth,
%   colspec = {Q[573]Q[60]Q[308]},
%   hline{2} = {-}{},
% }
\begin{table}[ht]
    \centering
    \caption{The ten test sentences selected from the test on which the prompt was engineered.}
    \label{table:expert_phase1_stats}
    \begin{tabular}{p{8cm}p{2cm}p{3cm}}
    % \begin{tabularx}{\textwidth}{llX}
    
    \toprule
    \textbf{Sentence} & \textbf{NGO} & \textbf{Source}  \\    
    ‘I look at the hills: where does my help come from?     & CCA  & \href{https://christianclimateaction.org/2021/06/11/solo-cycle-for-the-climate-crisis-day-4/}{CCA June 2021} \\
    
    At least four of St Paul’s letters are traditionally called the ‘prison epistles’: Ephesians, Philippians, Colossians, and Philemon.                                       &CCA          & \href{https://christianclimateaction.org/2021/08/26/fr-martin-newell-following-christ-to-prison-pt-1/}{CCA August 2021} \\
    {As I mentioned above, the apostles and disciples are imprisoned a number of times in the Book of Acts.}                                                            & CCA          & \href{https://christianclimateaction.org/2021/08/26/fr-martin-newell-following-christ-to-prison-pt-1/}{CCA August 2021} \\
    It was of course from prison that Dietrich Bonhoeffer penned those electrifying words in his ‘Letters and Papers from Prison’.                                             &CCA          & \href{https://christianclimateaction.org/2021/08/26/fr-martin-newell-following-christ-to-prison-pt-1/}{CCA August 2021} \\
    At least some Victorian era prisons were deliberately designed along similar lines to monasteries – it’s no accident that, like monasteries, prisons have cells, and ‘regular hours’ for meals. & CCA          & \href{https://christianclimateaction.org/2021/08/26/fr-martin-newell-following-christ-to-prison-pt-1/}{CCA August 2021} \\
    For Our Sacred Earth; Aug 28, 2024; Climate change, driven by fossil fuels, threatens the well-being of people and planet alike.                                      &GreenFaith   & \href{https://greenfaith.org/press-release-body-soul-against-eacop-2/page/2/?et\_blog}{Green Faith blog}                 \\
    And we ask Senator Murkowski, and other US Congressional members, to join us in this movement to protect Mother Earth and say no to the Keystone XL pipeline.              &IEN          & \href{https://www.ienearth.org/?p=2075}{ienearth.org}                                                               \\
    Sanctity of nature; Sikhs cultivate an awareness and respect for the dignity of all life, human or otherwise.                                                           &ICSD         & \href{https://interfaithsustain.com/?p=15296}  {interfaithsustain}                                                        \\
    The element of water is therefore a primary link in the interdependence of humanity and nature, to be used is in a sustainable and fair way.                               &ICSD         & \href{https://interfaithsustain.com/?p=15296}{interfaithsustain}                                                          \\
    Based on the Windsor Statements.                                                                                                                                           &ICSD         & \href{https://interfaithsustain.com/?p=15318}{interfaithsustain}          \\                                               \bottomrule
    \end{tabular}
\end{table}

% \end{landscape}
% \end{appendices}
\clearpage
\newpage

\printbibliography

%\bibliography{referencelist.bib}
% \bibliographystyle{apa}
\end{document}